\documentclass[letterpaper, 10 pt, journal, twoside]{IEEEtran}  

\IEEEoverridecommandlockouts                              

\usepackage{amsmath} 
\usepackage{amssymb}  
\usepackage{xcolor}
\usepackage{graphicx} %
\graphicspath{{images/}} %
\usepackage{subfig}
\usepackage{hyperref}
\usepackage{eurosym}
\usepackage{xargs}
\usepackage[colorinlistoftodos,prependcaption,textsize=footnotesize]{todonotes}
\usepackage[noadjust]{cite}

\usepackage[inline]{enumitem}

\usepackage[per-mode=symbol]{siunitx}            
\newcommand{\seq}{\,{=}\,} 

\newcommandx{\ludo}[2][1=]{\todo[inline,linecolor=blue!20,backgroundcolor=blue!20,bordercolor=blue,#1]{#2}}
\newcommandx{\alex}[2][1=]{\todo[inline,linecolor=blue!20,backgroundcolor=orange!20,bordercolor=orange,#1]{#2}}

\title{An Open Torque-Controlled Modular Robot Architecture for Legged Locomotion Research}

\author{Felix Grimminger$^{1}$%
, Avadesh Meduri$^{1,3}$%
, Majid Khadiv$^{1}$%
, Julian Viereck$^{1,3}$%
, Manuel W\"uthrich$^{1}$%
\\
Maximilien Naveau$^{1}$%
, Vincent Berenz$^{1}$
, Steve Heim$^{2}$%
, Felix Widmaier$^{1}$%
, Thomas Flayols$^{4}$\\
Jonathan Fiene$^{2}$%
, Alexander Badri-Spr\"owitz$^{2}$%
 and Ludovic Righetti$^{1,3}$
\thanks{This work was supported by New York University, Max-Planck Institute for Intelligent Systems' Grassroots projects, the European Union’s Horizon 2020 research and innovation program (grant agreement 780684 and European Research Council’s grant 637935), the National Science Foundation (CMMI-1825993), a Google Faculty Research Award, and an Independent Max Planck Researcher Grant.
We thank Joel Bessekon Akpo for his help with the motor driver testing and Joshi Walzog for his help with the foot sensor circuit board layout.}
\thanks{$^{1}$ Max Planck Institute for Intelligent Systems, 72076 T\"ubingen, Germany. Email \tt \footnotesize first.lastname@tuebingen.mpg.de}
\thanks{$^{2}$ Max Planck Institute for Intelligent Systems, 70569 Stuttgart, Germany. Email \tt \footnotesize lastname@is.mpg.de}
\thanks{$^{3}$ Tandon School of Engineering, New York University, Brooklyn, USA. Email \tt \footnotesize first.lastname@nyu.edu}
\thanks{$^{4}$ Laboratory for Analysis and Architecture of Systems, Centre National pour la Recherche Scientifique, Toulouse, France. Email \tt \footnotesize thomas.flayols@laas.fr}
}

\begin{document}

\maketitle

\begin{abstract}
We present a new open-source torque-controlled
legged robot system, with a low-cost and low-complexity actuator module at its core.
It consists of a high-torque brushless DC motor and a low-gear-ratio transmission suitable for impedance and force control.
%
%
We also present a novel foot contact sensor suitable for legged locomotion with hard impacts.
A 2.2 kg quadruped robot with a large range of motion is assembled from eight identical actuator modules and four lower legs with foot contact sensors.
%
%
Leveraging standard plastic 3D printing and off-the-shelf parts results in a lightweight and inexpensive robot, allowing for rapid distribution and duplication within the research community.
%
%
We systematically characterize
the achieved impedance at the foot in both static and dynamic scenarios, and measure a maximum dimensionless leg stiffness of \num{10.8} without active damping, which is comparable to the leg stiffness of a running human.
Finally, to demonstrate the capabilities of the quadruped, we present a novel
controller which combines feedforward contact forces computed from a kino-dynamic optimizer with
impedance control of the center of mass and base orientation.
The controller can regulate complex motions while being robust to environmental uncertainty.
\end{abstract}


\section{Introduction}
\IEEEPARstart{I}{t} is often difficult to test advanced control and learning algorithms for legged robots without significant hardware development efforts or maintenance costs. Present-day hardware is often mechanically complex and costly, different robot systems are hard to compare to one another, and many systems are not commercially available.
%
%
\begin{figure}
    \centering
    \includegraphics[width=0.45\textwidth]{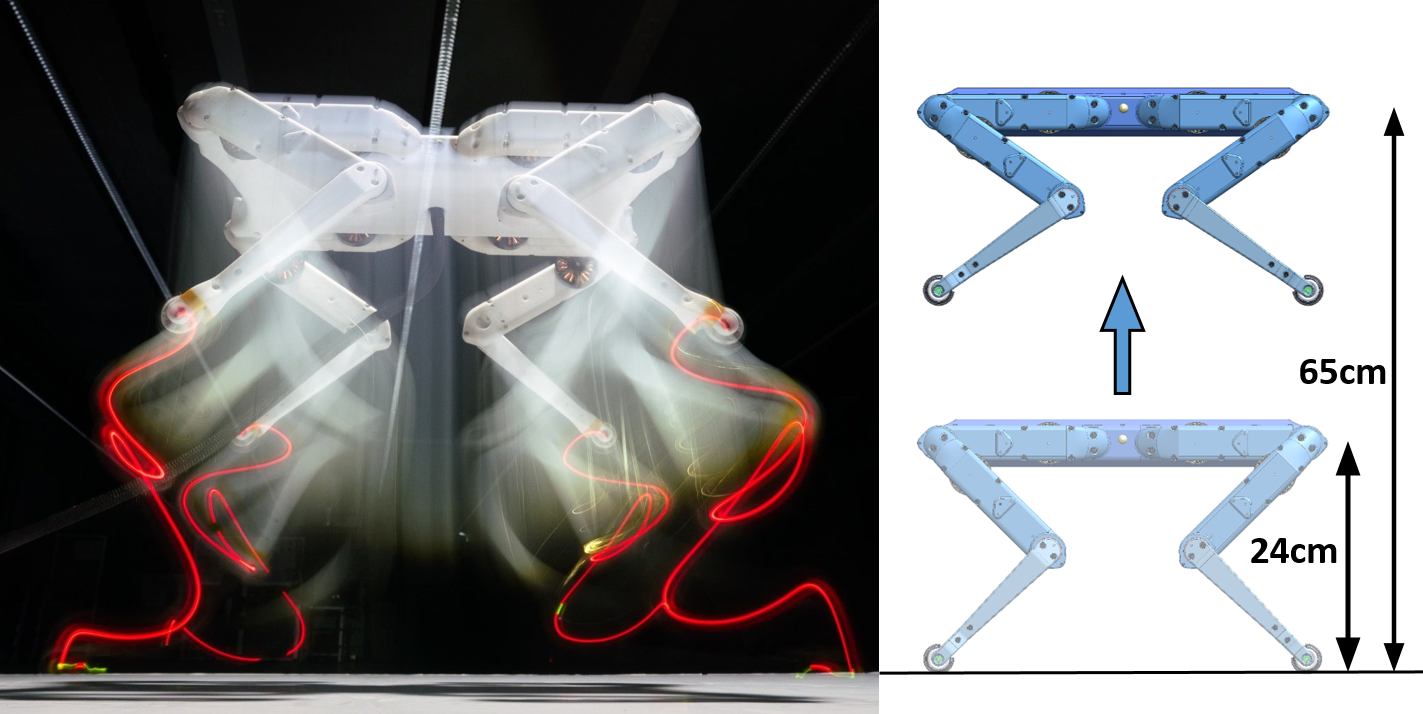}
    \caption{Quadruped robot Solo jumping: the maximum jump height at the time of writing was \SI{65}{cm} from a standing height of \SI{24}{cm}. 
    (Photo by W. Scheible)}
    \label{fig: solo quadruped}
\end{figure}
%
%
%
To support rapid and broad progress in academic research, we believe that hardware, firmware, and middle-ware must become inexpensive and relatively easy to reproduce and implement. Open-source blueprints of low-cost legged robot platforms like Oncilla~\cite{sprowitz_oncilla_2018}, Stanford-doggo~\cite{kau2019stanford}, and ours allow researchers to test and develop their ideas on a variety of robotic platforms with different morphologies. In addition, the performance and capabilities of robots produced from open-source blueprints can be directly compared across laboratories.

%

To the best of our knowledge Stanford-doggo is the only other open-source torque-controlled legged robot platform. Most of its parts are waterjet cut and relatively easy to produce. 
Other open-source (hobby) projects mostly use
position controlled servo motors, limiting their usage for advanced control and learning algorithms.
Complex machining is typical in quadruped robots exhibiting high performance force control such as HyQ \cite{semini_towards_2015}, Anymal \cite{hutter2016anymal}
or the MIT Cheetah \cite{wensing_proprioceptive_2017}.
Numerous legged robots such as the Oncilla and Cheetah-cub \cite{sprowitz_towards_2013} incorporate mechanical compliance in parallel to the leg actuators to react intrinsically and immediately to external perturbations; however, complicated mechanisms are required  to effectively alter joint stiffness~\cite{hurst_physically_2005}.  

For an open-source legged robot to be successful, it is necessary to minimize the number of parts requiring precision machining, thereby favoring off-the-shelf components over sophisticated custom actuation solutions.  To achieve this goal, we leverage recent advances in inexpensive plastic 3D printing and high-performance brushless DC motors (BLDC) which are now widely available as off-the-shelf components.  Furthermore, we can take advantage of the improvements driven by the mobile-device market, including affordable miniature sensors, low power and high-performance micro-controllers, and advanced battery technologies.

%



Lightweight, inexpensive yet robust robots are particularly relevant when testing advanced algorithms for dynamic locomotion \cite{carpentier2018multicontact,ponton2018time,mordatch2012discovery}. 
Indeed, ease of operation and collaborative open-source development can accelerate testing cycles. 
Lightweight robots require no cranes or guiding structures and can be operated by a single researcher in smaller laboratory spaces.
They can also significantly reduce the time and cost of repair and maintenance.
These features become especially important when testing learning algorithms directly on real hardware \cite{viereck2018learning,Tan-RSS-18} where a safe platform is required to explore various control patterns.


Towards these goals, we present a novel, fully open-source, modular force-controlled leg architecture for dynamic legged robot research.
This paper presents five main contributions:
1) a novel lightweight, low-complexity, 
torque-controlled actuator module suitable for impedance and force control  based on off-the-shelf components and 3D printed parts, 
2) an under 10g-lightweight foot contact sensor, capable of detecting contact reliably in all directions, suitable for robots withstanding hard impacts,
3) one complete leg constructed from two actuator modules and the foot sensor with controlled dimensionless impedance in the range of human running,
4) a quadruped robot, Solo, assembled from four legs with a total mass of only \SI{2.2}{kg},
and 5) a torque-controller for tracking full-body motions computed with a kino-dynamic
motion optimizer \cite{herzog2016structured, ponton2018time}, 
demonstrating for the first time 
the execution of such motions on real robots under moderate environmental uncertainty.
All designs are open-sourced, including mechanical drawings, electronic circuits, and control software \cite{opensourcelink}.

\begin{figure*}[!th]
\centering
\subfloat[Actuator module\label{subfig-1:dummy0}]{%
\includegraphics[height=.21\textwidth]{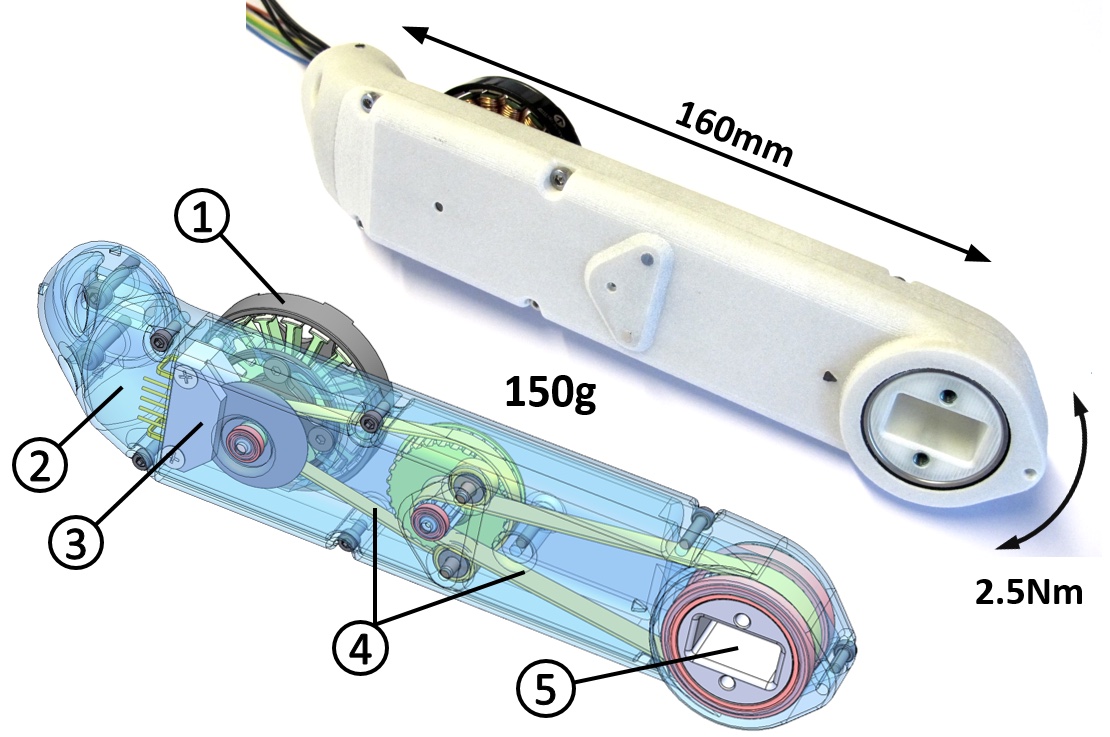}
}
\subfloat[Component overview\label{subfig-1:dummy1}]{%
\includegraphics[height=.21\textwidth]{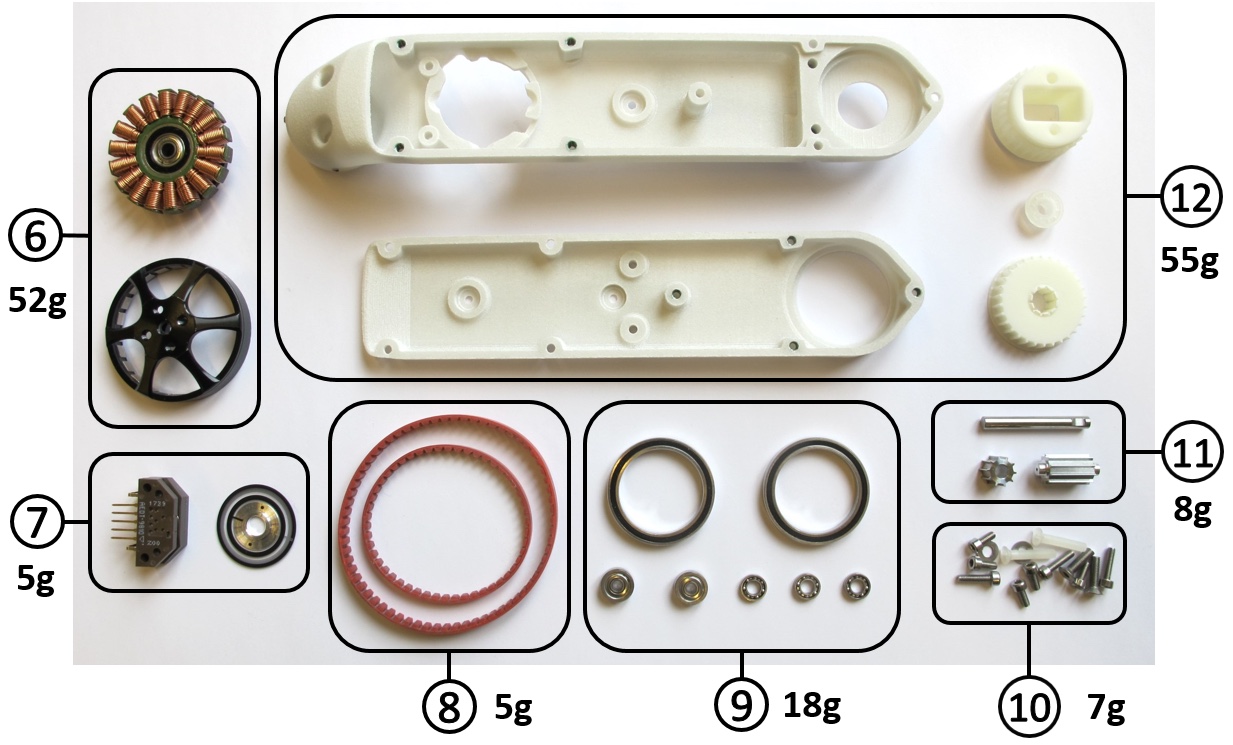}
}
\subfloat[Lower leg component overview\label{subfig-1:dummy2}]{%
\includegraphics[height=.21\textwidth]{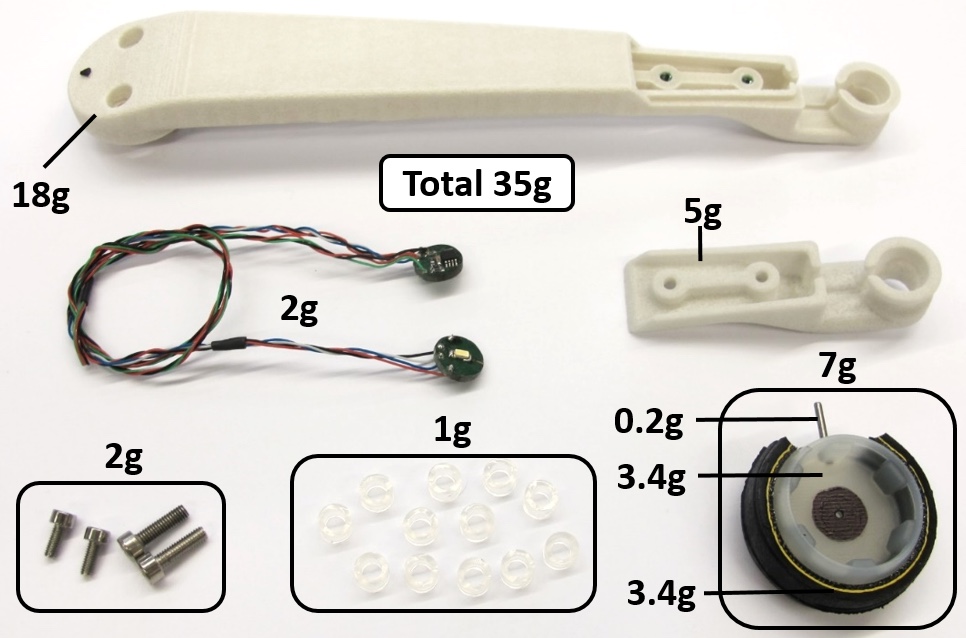}
}
\caption{%
    Brushless actuator module
    \textbf{(a)} 
    assembled, and
    \textbf{(b)}
    individual parts.
    BLDC motor~{\small\textcircled{\tiny 1}}, 
    two-part 3D printed shell structure~{\small\textcircled{\tiny 2}}, 
    high resolution encoder~{\small\textcircled{\tiny 3}},
    timing belts~{\small\textcircled{\tiny 4}},
    and output shaft~{\small\textcircled{\tiny 5}}.
    Brushless motor {\small\textcircled{\tiny 6}},
    optical encoder {\small\textcircled{\tiny 7}},
    timing belts {\small\textcircled{\tiny 8}},
    bearings {\small\textcircled{\tiny 9}},
    fasteners {\small\textcircled{\tiny 10}},
    machined parts {\small\textcircled{\tiny 11}}
    and 3D printed parts {\small\textcircled{\tiny 12}}.
    With the exception of {\small\textcircled{\tiny 11}}, all parts are either off-the-shelf, or printable on a regular 3D printer. The motor shaft and the pulleys {\small\textcircled{\tiny 11}} can be machined from stock material.
    \textbf{(c)} 
    Lower leg and foot contact switch components.
    }
\label{fig:2-dof-leg} 
\vspace{-0.2cm}
\end{figure*}
%
%
\section{PLATFORM AND ROBOT OVERVIEW}
This section details the actuator and contact sensor concepts
for our modular leg design leading to a two-DOF leg and the full quadruped robot, Solo.
%
%
%
\subsection{Actuator Concept}
\subsubsection{Brushless Actuator Module}
%
The actuator module  is shown in Figure\,\ref{fig:2-dof-leg}a.
It consists of a brushless motor (T-Motor Antigravity 4004, 300KV), a 9:1 dual-stage timing belt transmission (Conti Synchroflex AT3 GEN III), a high-resolution optical encoder (Avago AEDM 5810) and a 5000 count-per-revolution code wheel mounted directly on the motor shaft.
Everything is contained inside the lightweight, 3D printed shell, with no exposed wires.
The low transmission ratio enables reasonable peak torques and high velocity at the joint. 
Importantly, the design ensures sufficient transparency to enable accurate torque control through motor current measurements alone.
The module weighs \SI{150}{\gram} for a segment length of \SI{160}{\milli\meter}. 
The assembly of the module is simple and requires very few components, as can be seen in Figure \ref{fig:2-dof-leg}b. 
All components are either available off-the-shelf or can be 3D printed except for the motor shaft and the pulleys, which need to be machined from stock material.
Desired joint torques are converted into desired motor current using the relationship $\tau_{joint} = k_i \, i \, N$  where $k_i=\SI{0.025}{\newton\meter\per\ampere}$  and the gear reduction $N=9$ which leads to $\tau_{joint} = 0.225 \, i$.
The actuator can output $\tau_{max}=\SI{2.7}{N.m}$ joint torque at \SI{12}{\ampere}.

%
%
\subsubsection{Electronics}
%
%
The experiments presented in the paper use off-the-shelf TI micro-controller evaluation boards equipped with two BLDC booster cards each (Fig.\,\ref{fig:electronics comparison} {\large \textcircled{\footnotesize 13}}). 
They are capable of Field Oriented Control (FOC), and execute dual motor torque control at \SI{10}{\kilo\hertz} (Fig.\,\ref{fig: system_overview}).
To reduce the electronics footprint, we miniaturized the TI motor driver electronics, reducing the mass by a factor of six and volume by a factor of ten, as shown in Fig.\,\ref{fig:electronics comparison}.
The resulting open-source MPI Micro-Driver electronics {\large \textcircled{\footnotesize 14}} consist of a Texas Instruments micro-controller (TMS320F28069M) and two brushless motor driver chips (DRV8305) on a single six-layer printed circuit board.  The board includes a JTAG port for programming, CAN and SPI ports for control communications, and operates at motor voltages up to $\SI{40}{\volt}$.
%
%

\subsection{Foot Contact Sensor} 
\begin{figure*}[thb!]
    \centering
    \subfloat[Brushless system overview\label{fig: system_overview}]{%
\includegraphics[height=.23\textwidth]{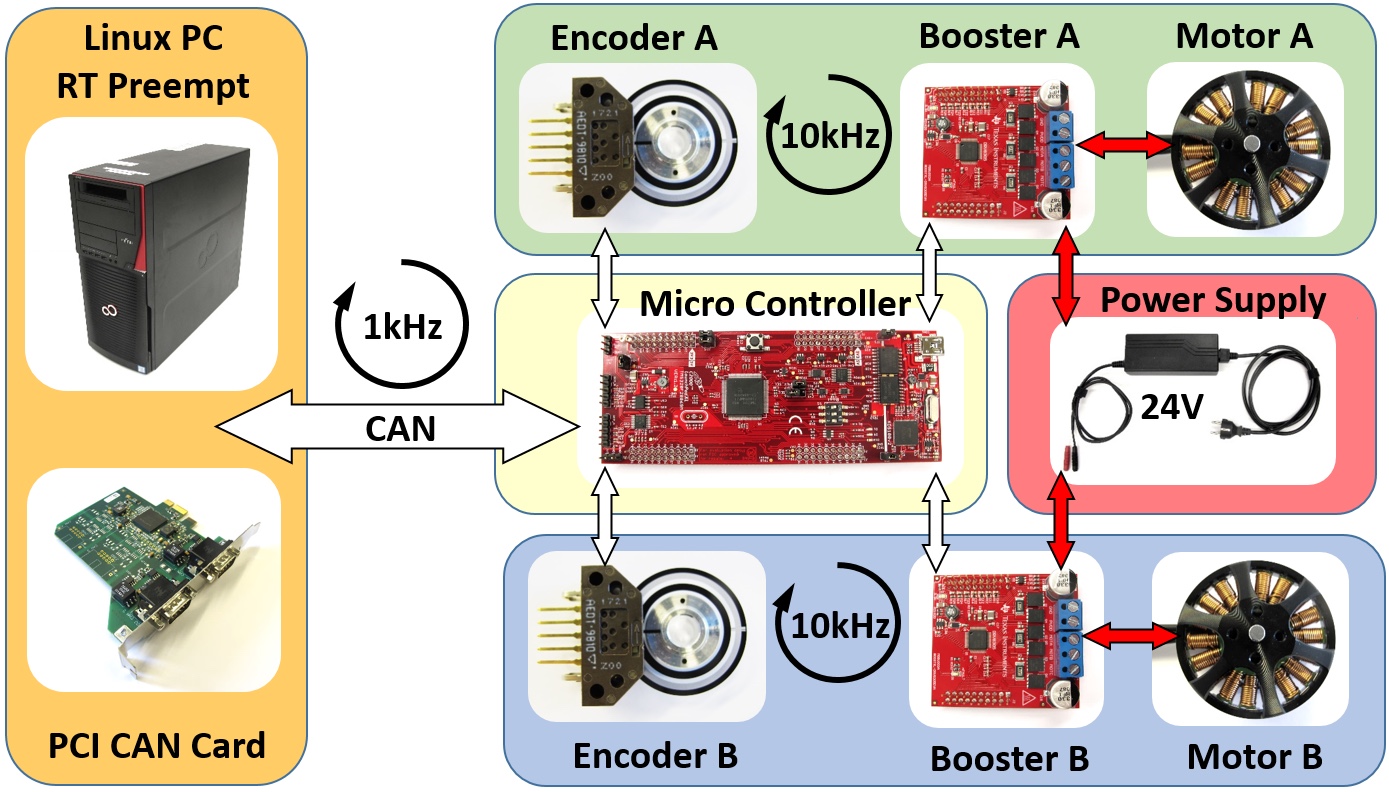}
}
\subfloat[Brushless motor driver boards\label{fig:electronics comparison}]
{\includegraphics[height=.23\textwidth]{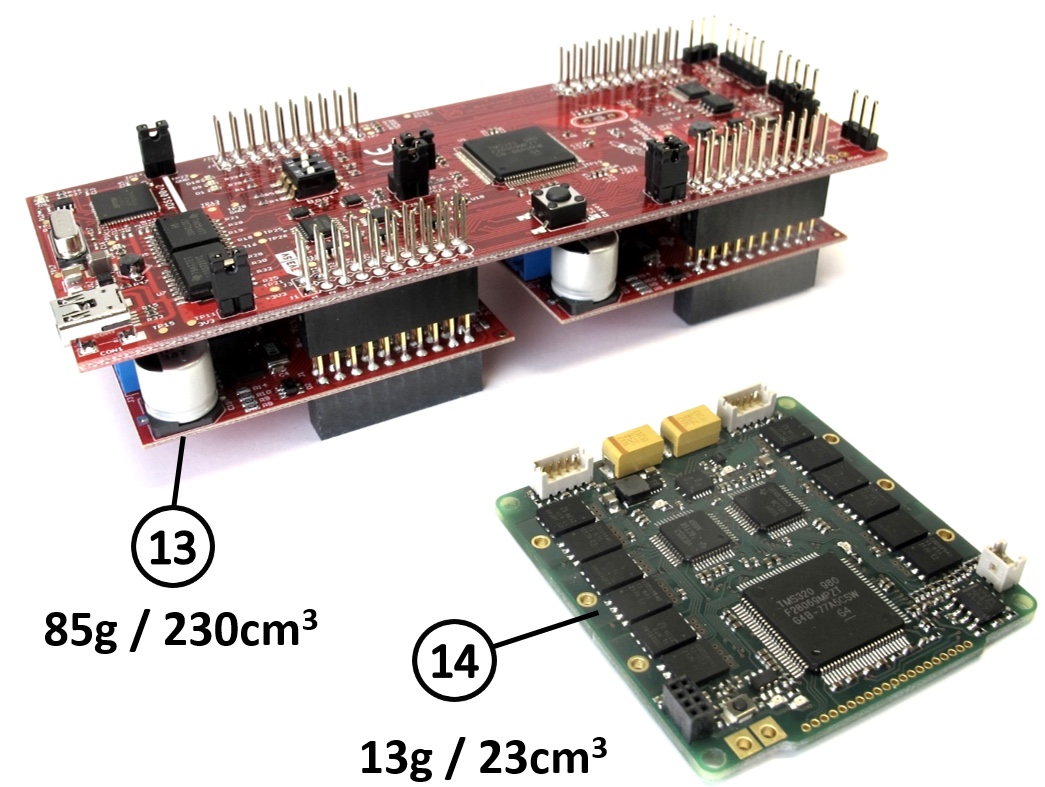}
}
\subfloat[Foot contact switch\label{fig:foot sensor}]
{\includegraphics[height=.23\textwidth]{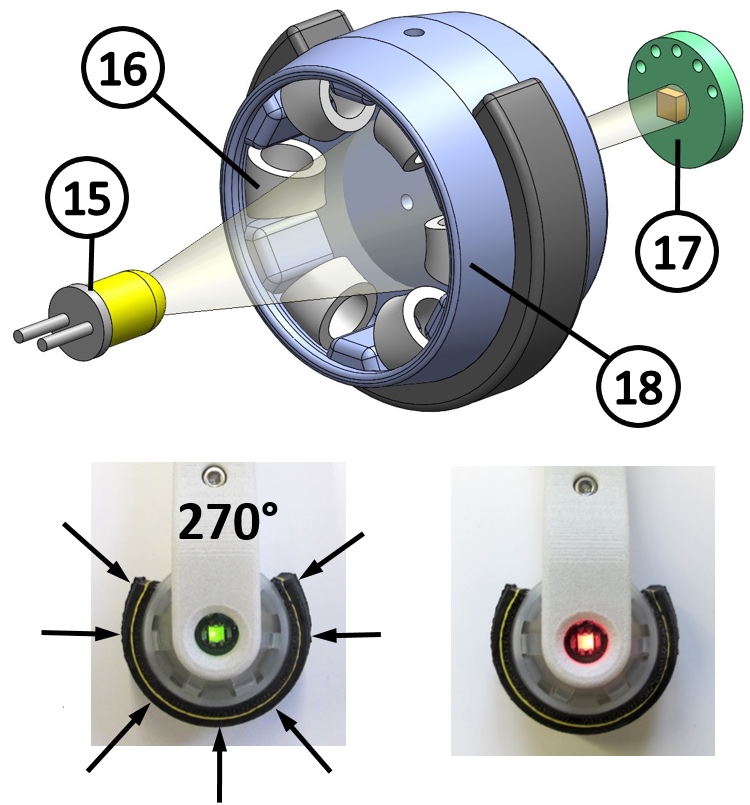}
}
\caption{%
\textbf{(a)}
A CAN bus connects the PC to a micro controller board (TI evaluation board). Two brushless DC motors are controlled at \SI{10}{\kilo\hertz} from each node. High resolution encoders provide motor shaft position feedback.  
\textbf{(b)}
We reduced an off-the-shelf TI Evaluation Board \textcircled{\tiny 13} into our MPI Micro-Driver electronics board \textcircled{\tiny 14} to control two brushless motors with feedback from two optical encoders.
\textbf{(c)} Our custom foot contact switch, activating at a \SI{270}{\degree} wide range of impact directions.
}
\vspace{-0.2cm}
\end{figure*}
%
The foot contact sensor (Figure \ref{fig:foot sensor}) is designed for rapid contact detection at a low force threshold and can withstand substantial impacts.
Since the foot's point-of-contact is often unpredictable in rough terrain, we designed the switch to activate with a sensing range of \SI{270}{\degree}, ensuring
proper contact detection for a wide range of robot configurations on complex terrains.
We implemented a mechanism based on a light-emitting diode {\large \textcircled{\footnotesize 15}} and a light sensor {\large \textcircled{\footnotesize 17}}. 
Both are separated by a spring-loaded, mechanical aperture {\large \textcircled{\footnotesize 18}} with a diameter of \SI{1.5}{\milli\meter}. The sensitivity of the contact switch can be adjusted by changing the diameter, length, and/or the number of elastic silicone elements  {\large \textcircled{\footnotesize 16}}.
External forces shift the aperture up to \SI{2}{\milli\meter} and generate an analog output signal between \SI{0}{\volt} and \SI{3}{\volt} that is read by an analog-to-digital converter on the micro-controller.
The foot and lower leg structures are 3D printed from plastic (Fig.\,\ref{fig:2-dof-leg}c).
The foot contact switch weighs \SI{10}{\gram} and triggers reliably at  \SI{3}{\newton} within \SI{3}{\milli\second} of contact.
The sensor is simple to assemble, has high sensitivity, low response time, and can withstand high impact loads. This makes it
suitable to detect contacts during dynamic locomotion tasks.

\subsection{2-DOF Leg and Quadruped Robot Solo}
\begin{figure*}[thb!]
\centering
\subfloat[2-DOF leg\label{fig:brushless leg}]{%
\includegraphics[height=.235\textwidth]{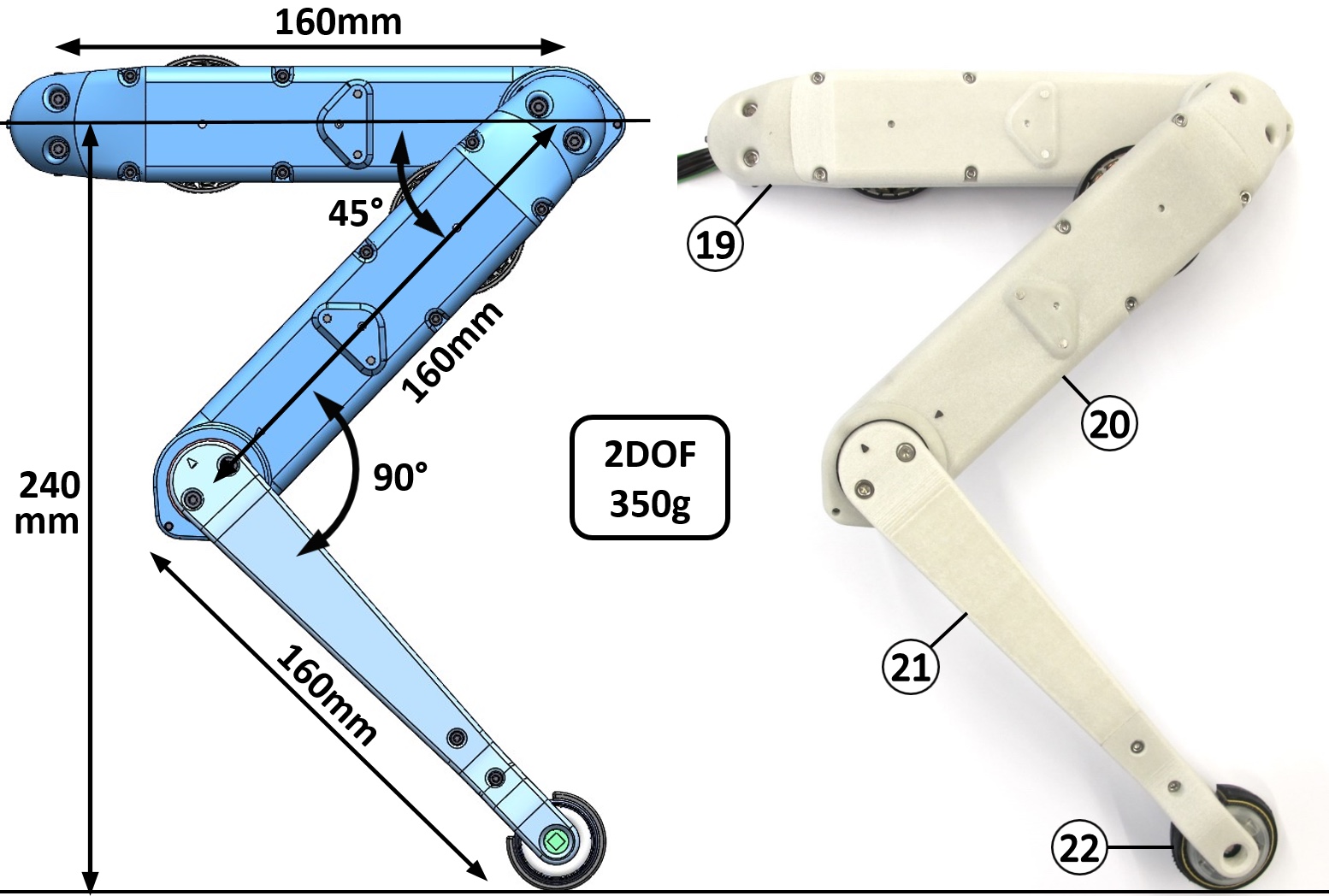}
}
\subfloat[Impedance control\label{fig:impedance}]{%
\includegraphics[height=.235\textwidth]{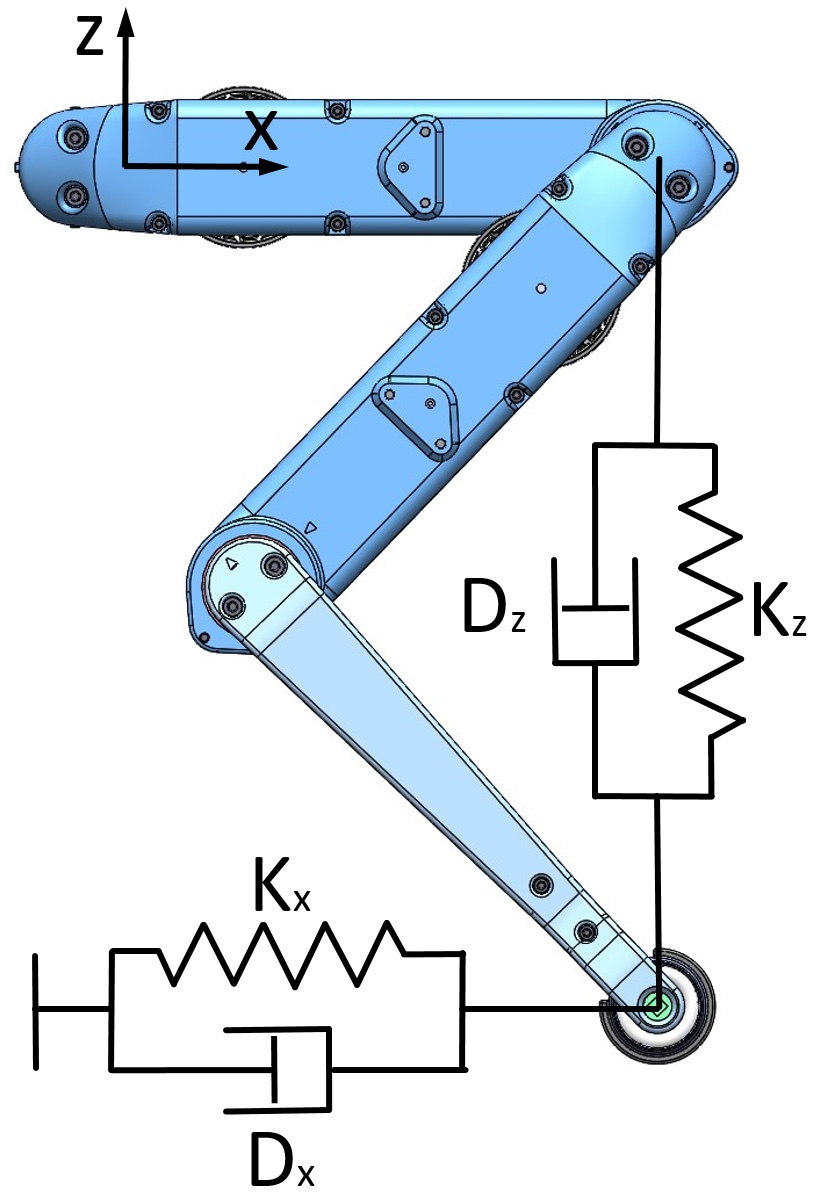}
}
\subfloat[The 8-DOF Quadruped robot 'Solo'\label{fig:quadruped cad}]{%
\includegraphics[height=.22\textwidth]{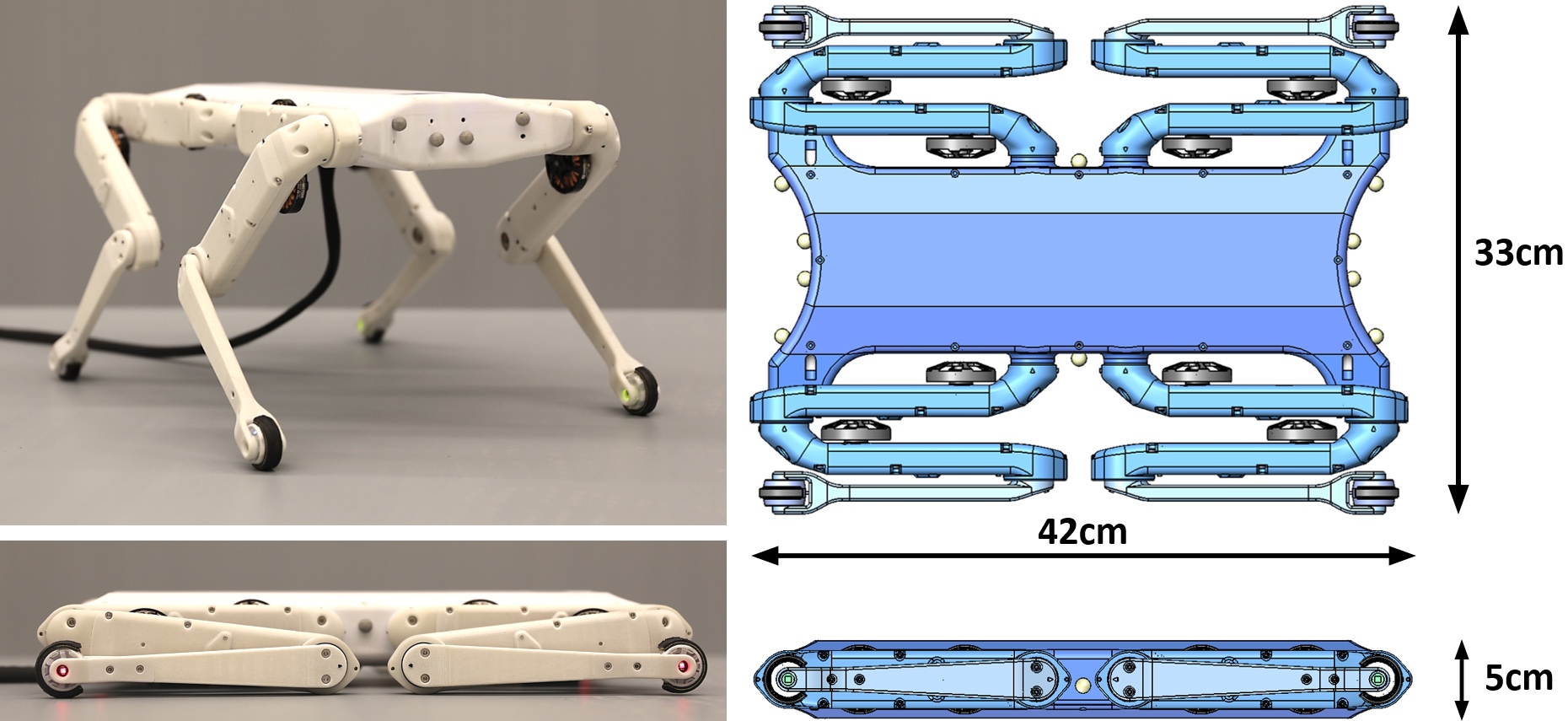}
}
\caption{%
\textbf{(a)}
Assembly of two brushless actuator modules hip \textcircled{\tiny 19}, and upper leg \textcircled{\tiny 20}, lower leg \textcircled{\tiny 21}, and foot contact switch \textcircled{\tiny 22}. At \SI{90}{\degree} knee angle, the 2-DOF leg stands \SI{0.24}{\meter} high, maximum hip height is \SI{0.34}{\meter}. 
\textbf{(b)}
Schematic presentation of impedance framework in Cartesian coordinates.
\textbf{(c)}
The \SI{2.2}{\kilo\gram} quadruped robot can fold into a \SI{5}{\centi\meter} flat structure.
}
\end{figure*}

A single, 2-DOF leg (Fig. \ref{fig:brushless leg}) is composed of two identical brushless actuator modules (hip {\large \textcircled{\footnotesize 19}}, upper leg {\large \textcircled{\footnotesize 20}}), and a lower leg {\large \textcircled{\footnotesize 21}}. The foot {\large \textcircled{\footnotesize 22}} is mounted distally, at the end of the lower leg. 
All joints are multi-turn capable. 
Cable routing between hollow segments limits rotations to about three turns in each direction.
%
%

We assembled the quadruped robot Solo from four identical legs and a 3D printed body structure (Figure \ref{fig:quadruped cad}). The robot's eight DOF are mounted to enable movements in the sagittal plane.
The trunk houses the electronics for controlling 8 BLDC motors. Solo is currently tethered for data transmission and external power. The robot weighs \SI{2.2}{\kilo\gram} with a standing hip height of approximately \SI{24}{\centi\meter} (maximum hip height of \SI{34}{\centi\meter}), \SI{42}{\centi\meter} body length, and \SI{33}{\centi\meter} width. 
The robot can fold down to \SI{5}{\centi\meter} in height (Fig.\,\ref{fig:quadruped cad}). It is also symmetric about all three axes.
%
\subsection{Communication and Control Software}
While a CAN port for wired communication is available
on both the TI and custom electronic boards, we additionally designed a lightweight master board gathering all communications between the robot motor drivers and the off-board control computer. This board is based on an ESP32, a dual core \SI{240}{\mega\hertz} Xtensa LX6 SoC with Wifi and Ethernet interface.  Utilizing the upstream SPI port on the MPI Micro-Driver, dual motor state and control commands data are exchanged in a \SI{35}{\micro\second} long single bidirectional SPI transfer. The master board is able to control up to 8 dual motor drivers and connect to other sensors such as IMU and battery monitor. The board can be connected with the control computer via wired (Ethernet) or wireless (WiFi) network interfaces. The connection is point to point using raw MAC layer packets and vendor specific action frames for real time communication. We have achieved consistent \SI{200}{\micro\second} round trip time over 100 Mbit/s Ethernet, including driver latency.  The WiFi connection carries a round trip time of approximately \SI{1100}{\micro\second} on a free channel, with an average of about 4\% packet loss. The protocol always transmits the full state robot commands and sensors data, to be robust against the loss of packets over WiFi. We disabled WiFi acknowledgment packages by using multi-cast to ensure deterministic transfer time, and to give priority to new and more up-to-date data. With both interfaces, we can close a control loop with a 12-DOF setup at \SI{1}{\kilo\hertz}. 
This open-source master board design is being used in the newest versions of the robot together with the MPI Micro-Driver electronics.

Both the 2-DOF leg and the quadruped are remotely controlled by a PC running Ubuntu patched with RT-Preempt for real-time capabilities.
The sensing-control loop is performed at \SI{1}{\kilo\hertz}.
%
%
%
We implemented drivers to interface the electronics with the control PC, with options for CAN, Ethernet or WiFi communication.
A C++ software package provides an API to interface with several motor boards from the PC, with basic functionalities for position and force control. 
The API comes with Python bindings, enabling rapid prototyping. We use the  C++ interface to implement \SI{1}{\kilo\hertz} control loops.
Several demo programs to rapidly test multi-actuator control are provided in the open-source repository \cite{opensourcelink}.

\section{Experiments and Results}
In this section, we present experiments with the 2-DOF leg and the quadruped robot. 
We quantify the impedance regulation properties of the system, then
we present a controller to track motions optimized with a kino-dynamic optimizer
and demonstrate dynamic behaviors on the quadruped robot. 

\subsection{Impedance control of the 2-DOF leg}
\begin{figure}[hbt!]
    \centering
    \includegraphics[width=0.35\textwidth]{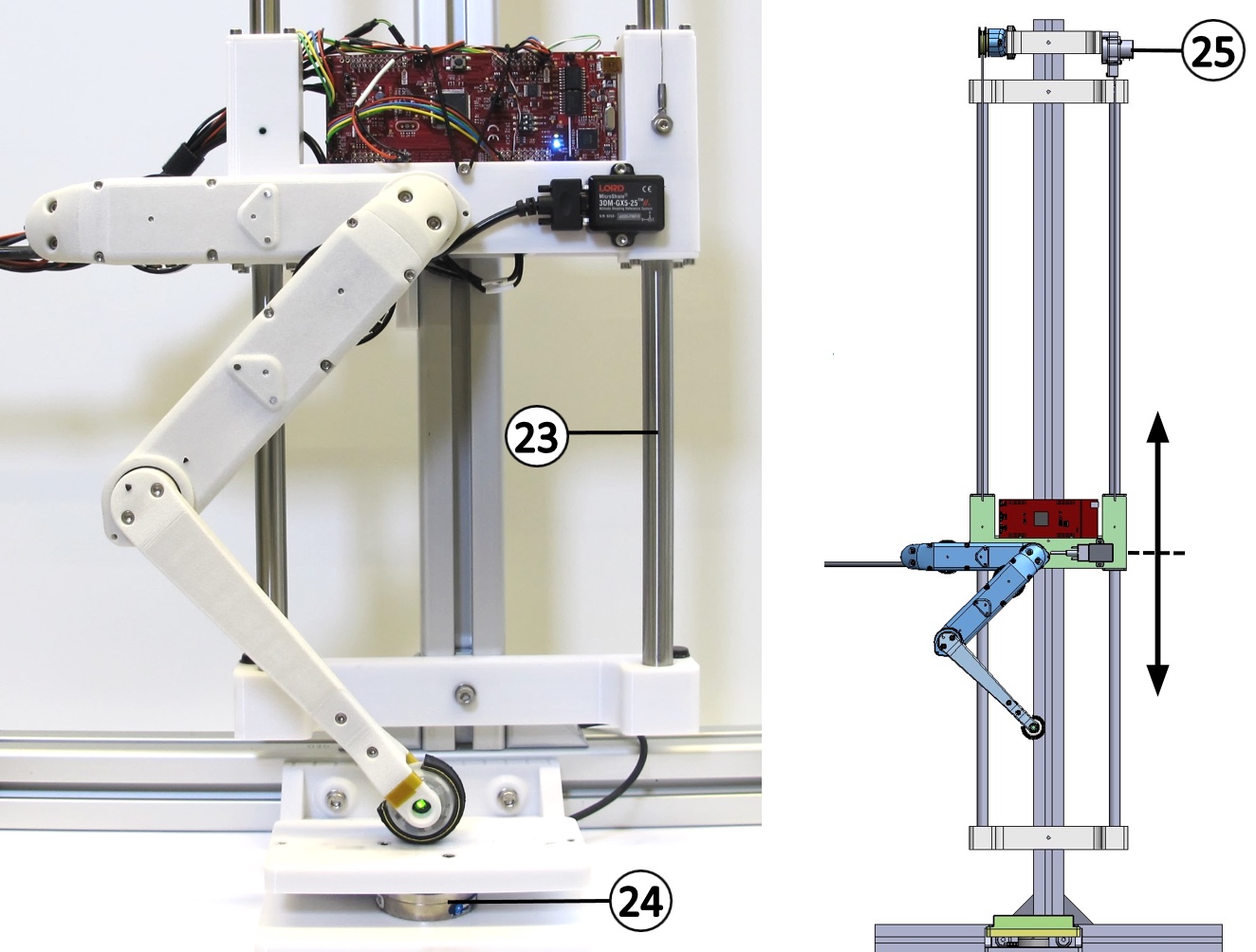}
    \caption{
    Leg test stand with linear guide {\textcircled{\tiny 23}}.
    A 6-axis ATI Mini40 force sensor {\textcircled{\tiny 24}} measured ground reaction forces.
    A string potentiometer measures leg height {\textcircled{\tiny 25}}. 
    }
    \label{fig:teststand}
\end{figure}

We characterize the effective impedance control capabilities of the leg by
measuring the range of stiffness that can be regulated in quasi-static and hard impact conditions.
We built a test stand (Fig. \ref{fig:teststand}) with instrumentation to characterize the leg's stiffness profile.
%
%
A Cartesian impedance controller (Fig. \ref{fig:impedance}) regulates the stiffness and damping
of the foot with respect to the hip $\boldsymbol{\tau} \seq \mathbf{J}^T \left( \mathbf{K} (\mathbf{x}_d - \mathbf{x}) - \mathbf{D} \dot{\mathbf{x}} \right)$, where $\mathbf{x}\in\mathbb{R}^2$ is the foot position with respect to the hip (leg length), $\mathbf{x}_d\in\mathbb{R}^2$ the spring setpoint, $\mathbf{J}$ the foot Jacobian,
$\mathbf{K}$ and $\mathbf{D}$ the desired leg stiffness and damping matrices and $\boldsymbol{\tau}\in\mathbb{R}^2$ the motor torques. 
Torque control is only based on the internal motor current and motor position measurements without any force feedback. We validate impedance control quality using external reference sensors on the test stand.


\subsubsection{Quasi-static experiment}
We systematically characterized the range of stiffnesses that can be regulated at the foot for quasi-static motions.
The robot stood in a rest position, and we slowly pushed on the leg to produce a deflection. 
We measured the ground reaction force and the leg length using external ground-truth sensors (force plate and string potentiometer). 
In this experiment, $\mathbf{D} = \mathbf{0}$, and we only use the desired stiffness.
We found we could regulate the range of desired stiffness between \SI{20}{\newton\per\meter} and \SI{360}{\newton\per\meter}, for slow motion. For larger stiffness, without damping, the leg became unstable.
Note that with small amounts of damping, the maximum stiffness can be increased further while avoiding unstable controller behavior (not shown here).

\begin{figure}[thb!]
    \centering
    \includegraphics[width=0.85\columnwidth,trim={1cm 8cm 1cm 8cm}]{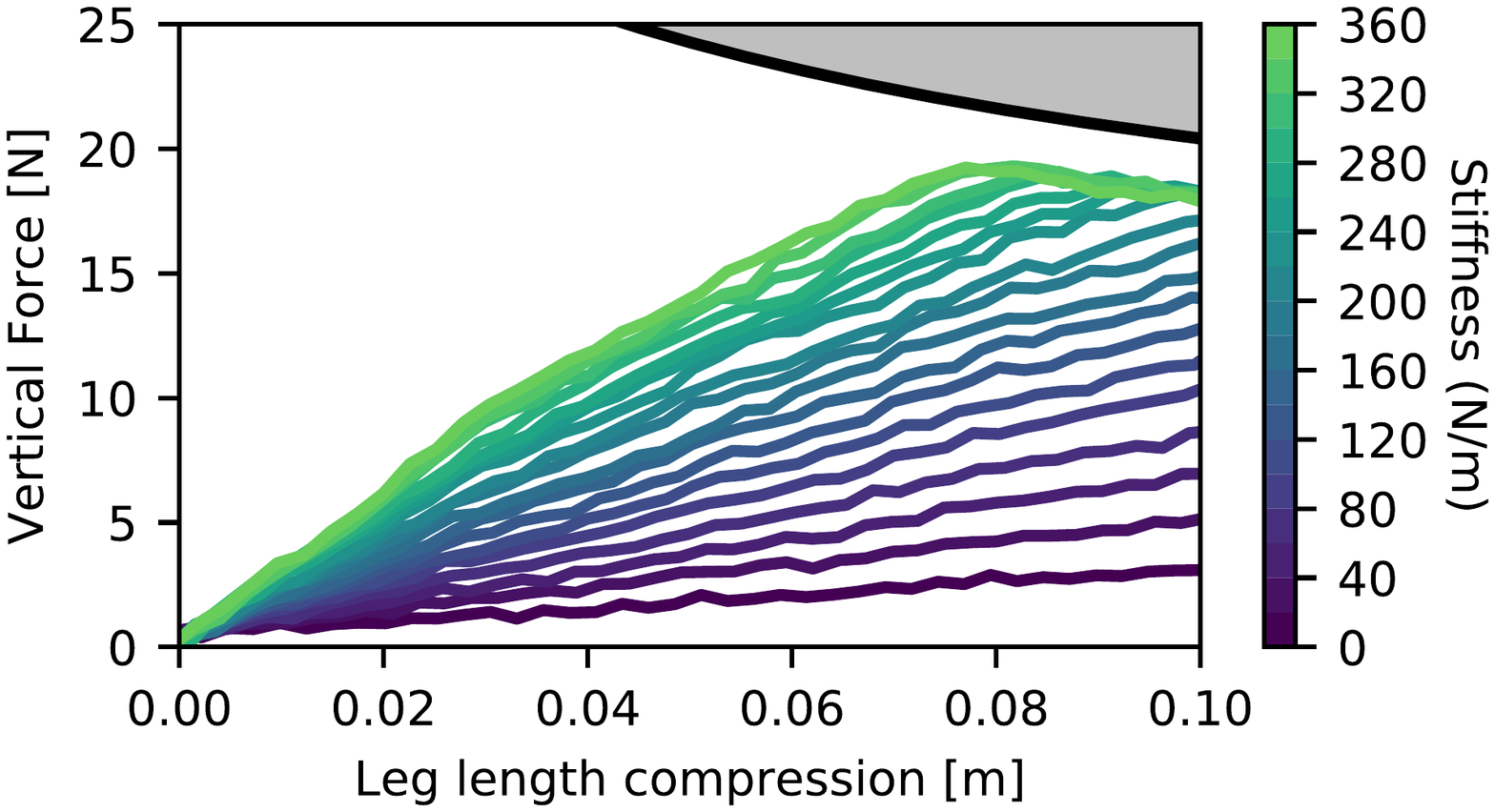}
    %
    \includegraphics[width=0.85\columnwidth]{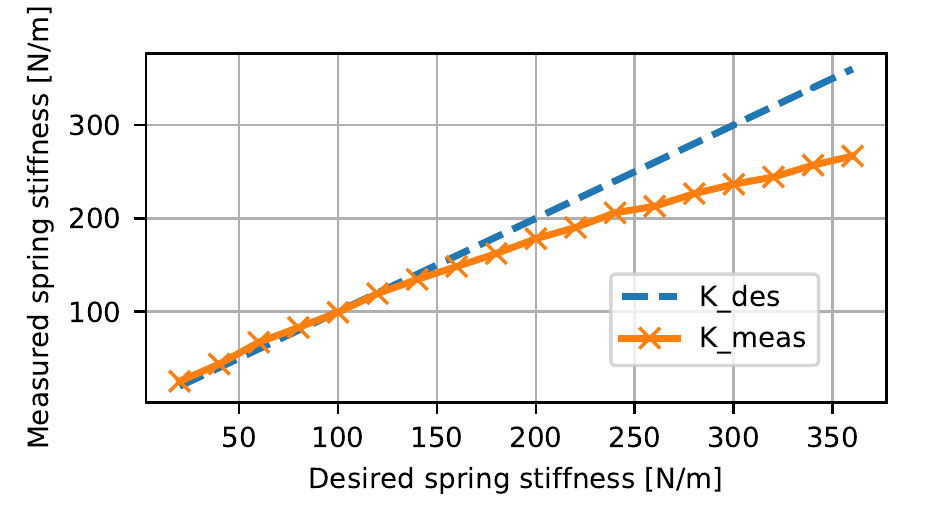}
    \caption{%
    Quasi-static experiment.
    Top: vertical ground reaction force versus leg compression, for desired stiffness ranging
    from \SI{20}{\newton\per\meter}
    to \SI{360}{\newton\per\meter} 
    in \SI{20}{\newton\per\meter} increments. 
The black asymptotically falling curve indicates the theoretical maximum leg force, calculated from the maximum knee torque and two-segment leg kinematics.
    Bottom: desired leg spring stiffness vs measured leg spring stiffness (from linear regression of the data shown in the top plot).
    }
    \label{fig:exp_A}
\end{figure}

Results of the experiment are shown in Figure \ref{fig:exp_A}.
We observe a close-to-linear relationship between vertical ground reaction force 
and vertical leg displacement for all desired stiffness values until actuator limits are reached (Fig. \ref{fig:exp_A}).
The maximum leg force (black line) is limited due to a combination of actuator torque limits (maximum applied current), and leg kinematics.
The linear relationship is independent of leg compression, suggesting that the linear impedance control
law works for a broad range of displacements (up to \SI{10}{\centi\meter} for cases below torque-saturation limits).

We computed the effective leg stiffness using linear regression, where we excluded data points in the saturation region. 
For commanded stiffness lower than \SI{150}{\newton\per\meter}, the measured leg stiffness matches the desired stiffness very well, while at higher desired stiffness, we observe lower measured stiffness. 
Without damping, the maximum measured stiffness was approximately \SI{266}{\newton\per\meter} for a commanded stiffness of \SI{360}{\newton\per\meter}.
The identification presented in Figure \ref{fig:exp_A} 
could also help choose a command to realize a reference stiffness.

The experiments demonstrate the ability of the robot to regulate leg stiffness with a simple
control law and without the need for torque sensing. Experimental data shows the linearity of the force-displacement relationship.
The difference in high stiffness regimes between actual and commanded stiffness is likely due to other dynamic effects
including friction, the flexibility of the transmission and error in joint position measurements (i.e., the encoders measure motor displacement, not joint motion). 

%
%
\subsubsection{Drop experiment} 
\begin{figure}[thb!]
\subfloat{
\includegraphics[width=.85\columnwidth]{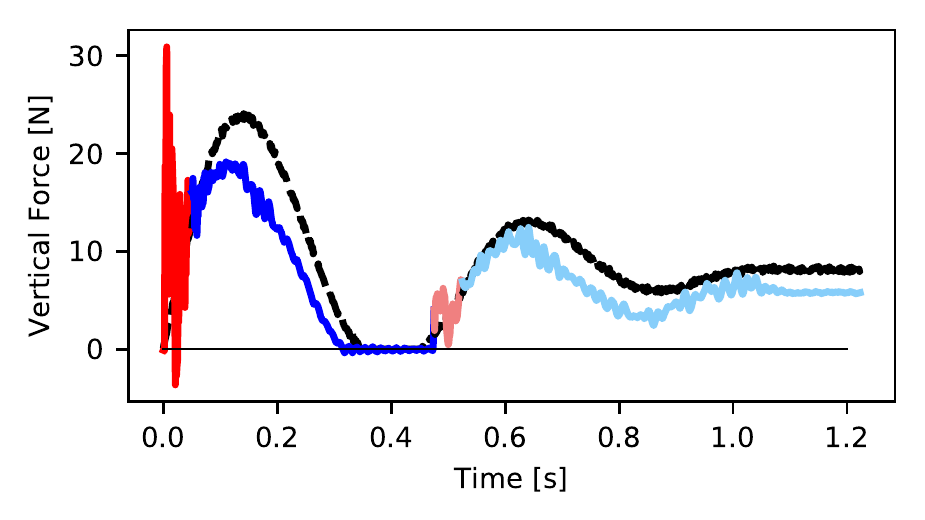}
}

\subfloat{
\includegraphics[width=.85\columnwidth]{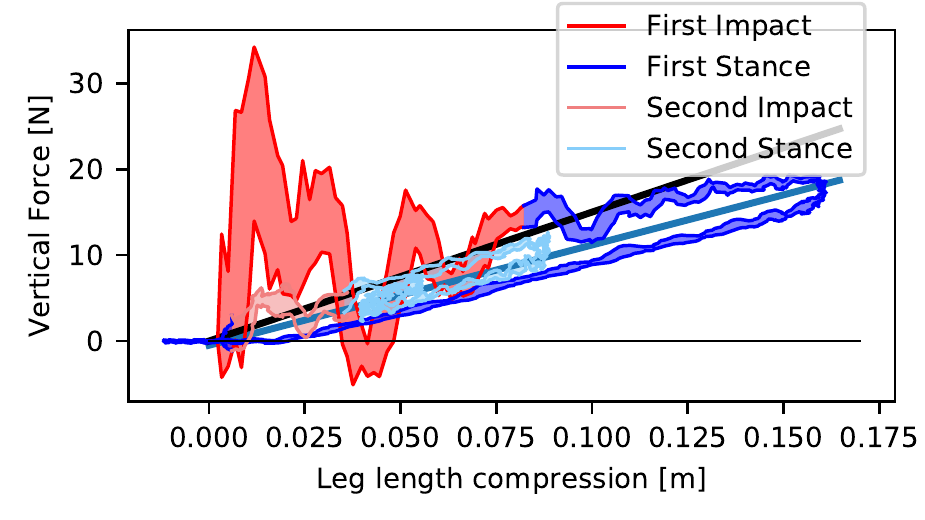}
}
\caption{%
Drop experiments repeated ten times:
the leg bounces twice (first bounce in red and blue, second bounce in light red and light blue). The black lines show the desired force response (desired stiffness multiplied by leg compression measured from the height sensor). Top: leg force as a function of time for one typical
experiment. Bottom: leg force as a function of leg compression (summary of all ten experiments).
The envelop is the average $\pm$ the standard deviation. The blue straight line is the regressed approximation of the stance data ($K=\SI{116}{\newton\per\meter}$ and $D=\SI{0.5}{\newton\second\per\meter}$).
}
\label{fig:dynamic_drop}
\end{figure}
The 2-DOF leg was dropped from a height of \SI{0.24}{\meter}, with a desired leg stiffness of \SI{150}{\newton\per\meter} and low damping
of \SI{0.5}{\newton\second\per\meter}.
This experiment shows the leg's impedance capabilities: producing simultaneous high torques and speeds.
Fig.\,\ref{fig:dynamic_drop} (top) shows a typical time evolution of the contact force.
The impact response of unsprung mass is visible as large, oscillating forces during the first \SI{50}{\milli\second} of the impact.
Frictional and deformation losses are visible by deviations from the ideal leg force: higher forces between touch-down and mid-stance, and lower forces between mid-stance and toe-off.
Losses lead to a lower second peak amplitude.
Leg forces settle at \SI{6}{\newton}, corresponding to the weight of the 2-DOF leg, vertical slider, and electronics.

Fig.\,\ref{fig:dynamic_drop} (bottom) shows hysteresis in the work space, and indicates friction and losses in structural system deformation. 
The hysteresis can be explained by Coulomb friction shifting forces above and below the desired force, depending on the desired direction of motion.
These hysteresis losses can be compensated with active control, however this was not the goal in this experiment. 
The low variance after impact shows good repeatability of the system. 
The linear relationship between leg compression and ground reaction forces remained.

\subsubsection{Jumping experiments}
We have already demonstrated the jumping capabilities of a preliminary version of the leg in \cite{viereck2018learning}.
Here we implement a simple periodic vertical motion using joint position controllers to generate jumping behavior.
The leg can jump approximately \SI{0.65}{\meter}, which is roughly twice its leg length and \num{2.7} times its resting leg length. 
The robot lands without damage. 
This demonstrates the ability of our system to generate dynamic motions.

\subsubsection{Contact sensor validation}
\begin{figure}
    \centering
    \includegraphics[width=0.45\textwidth]{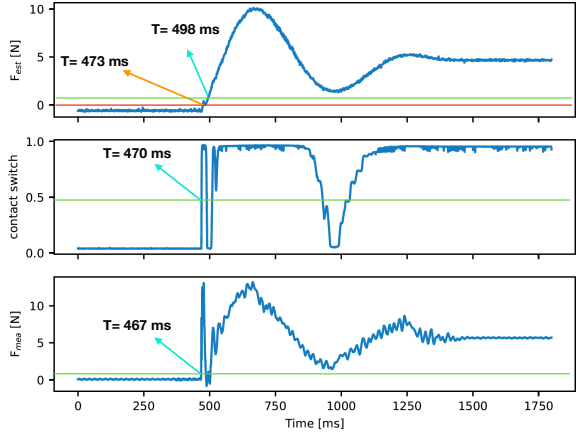}
    \caption{Drop test experiment. We notice a sharp contact detection with the contact sensor (middle) compared to force contact prediction from motor current (top). The bottom plot shows the ground truth from a high resolution force sensor.}
    \label{fig:contact_sensor}
\end{figure}

To  validate  the  performance  of  the  contact  sensor, we  performed  an  extensive  set  of  drop  tests  with  different  desired  leg  length ($l_{des} = \SI{20}{cm}$ and $l_{des} = \SI{30}{cm}$, close to singularity) and  impedance ($75$, $150$ and $\SI{300}{N/m}$) on  the  leg  test  stand.
We  compared  the  performance  of  the  contact sensor against estimated force at the foot using motor current
measurements  ($F=(J_a^T)^{-1} \tau$).
A threshold value is required for each sensing modality, to identify contact.
For the contact sensor, we set a threshold to 50 percent of the normalized value which resulted in a maximum delay of $\sim \SI{3}{ms}$ compared to the ground truth force detection from the test force sensor (Fig. \ref{fig:contact_sensor}).
Relying only on the joint torques estimated from motor currents, the threshold that ensures no false positives results in delays of $\sim \SI{31}{ms}$, which is significant for dynamic tasks where the entire contact duration is of the order of a few $\SI{100}{ms}$. Setting a lower threshold would lead to recurring false positive errors when changing leg length or impedance.
More advanced estimators using joint currents and a model of the full actuator drive system might improve detection at the cost of algorithmic complexity, while our contact sensor offers a simple, lightweight and robust solution.

\subsection{Dynamic behavior of the quadruped robot}
Here,  we demonstrate the dexterity and capabilities of the quadruped robot. Moreover, we present the first real-robot experiments using motions computed with a centroidal dynamics-based kino-dynamic planner \cite{herzog2016structured, ponton2018time}.
\begin{figure*}
    \centering
    \vspace{0.4cm}
    \includegraphics[width=0.85\textwidth]{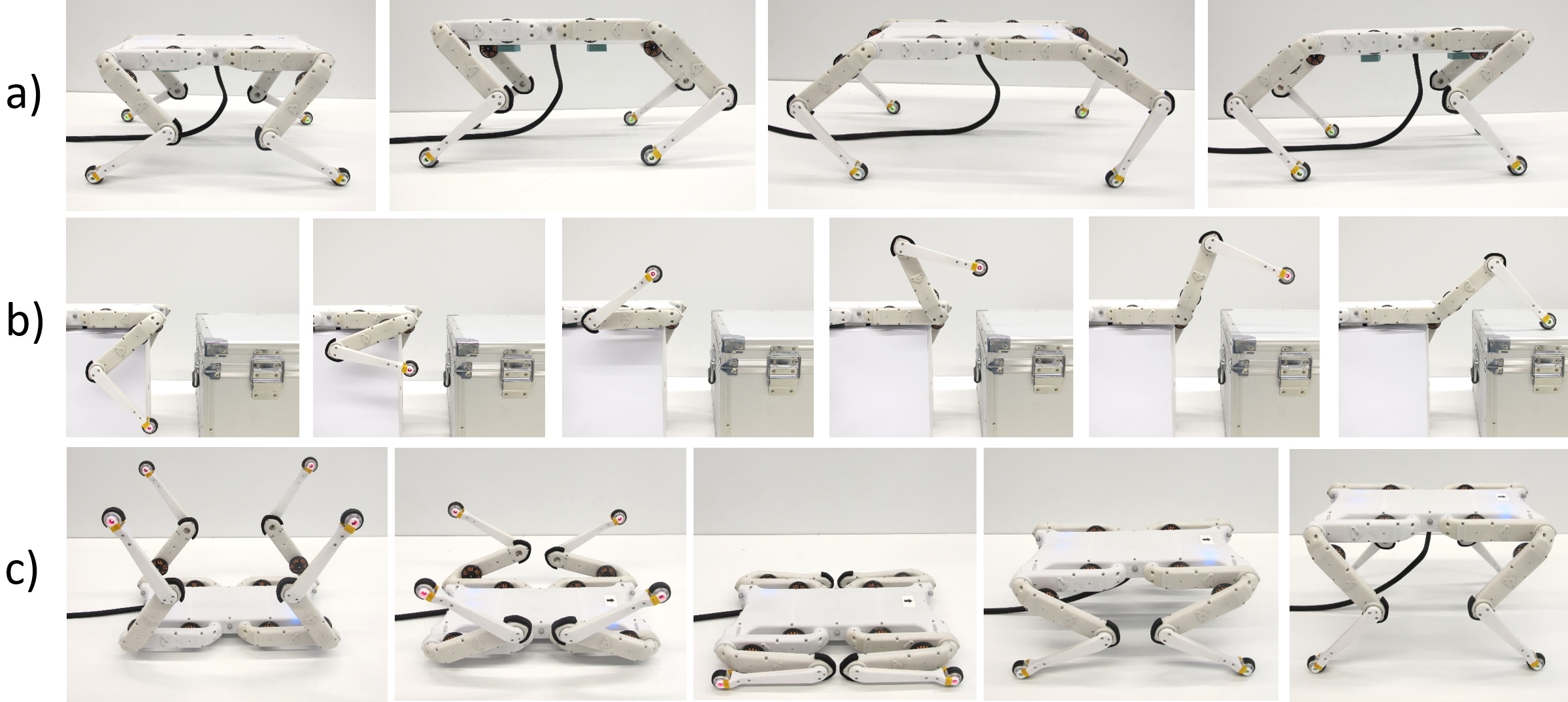}
    \caption{Example motion sequences: 
    a) Legs can switch between all the four knee configurations,
    b) with more than \SI{360}{\degree} hip joint rotation capability, and little space to navigate, legs can be rotated first backwards, and then onto a step,
    c) in case the robot falls onto its back, it can re-orient its legs, and stand up without rotating the trunk.
    }
    \label{fig:dexterity}
\end{figure*}
\subsubsection{Kino-dynamic motion optimizer and controller}
The motions are planned using the full body kino-dynamic optimizer proposed
in \cite{herzog2016structured}. The algorithm alternatively optimizes
1) the centroidal dynamics of the robot (i.e. its center of mass, linear and angular momentum) together with contact force trajectories
and 2) the kinematics of the full body. After several iterations, a consensus between
both optimization problems is reached, leading to locally optimal trajectories
consistent with the full robot dynamics. In our experiments, consensus was typically
achieved after two iterations.
The centroidal dynamics optimization problem is solved using the algorithm
proposed in \cite{ponton2018time}.

Similar to a passivity controller for multi-contact
balancing \cite{henze2016passivity}, our controller optimally distributes contact forces on the feet
to generate a sufficient wrench at the center of mass (CoM) to regulate reference CoM, angular momentum, and base orientation trajectories. In contrast to \cite{henze2016passivity}, we regulate the angular momentum of the robot and we add a low impedance controller for the motion of each foot, which is important
to handle hard impact dynamics.
The desired wrench $\mathbf{W}_{CoM}$ at the CoM is
\begin{equation}
    \mathbf{W}_{CoM} = \mathbf{W}_{CoM}^{ref} + \begin{bmatrix} 
    \mathbf{K}_{c} (\mathbf{x}_{c}^{ref} - \mathbf{x}_{c}) + \mathbf{D}_{c} (\dot{\mathbf{x}}_{c}^{ref} - \dot{\mathbf{x}}_{c}) \\
    \mathbf{K}_{b} (\mathbf{q}_{b}^{ref} \boxminus \mathbf{q}_{b}) + \mathbf{D}_{b}(\mathbf{k}^{ref} - \mathbf{k})
    \end{bmatrix}\nonumber
\end{equation}
where $\mathbf{x}_c$, $\mathbf{k}$ and $\mathbf{q}_b$ are the measured CoM position, angular momentum and base orientation (quaternion) respectively. The $^{ref}$ superscript denotes the reference trajectories from the motion optimizer, in particular, $\mathbf{W}_{CoM}^{ref}$
is the reference wrench at the CoM. The gain matrices
$\mathbf{K}_c$, $\mathbf{K}_b$, $\mathbf{D}_c$ and $\mathbf{D}_b$ are positive definite. 
In our experiments, we manually set these gains (diagonal matrices) but automatic designs such as in \cite{herzog2016momentum} could also be used.
The operator $\boxminus$
maps the rotation needed to correct for the orientation error between two quaternions into an angular velocity vector using the logarithm mapping between a Lie group and its Lie algebra.
This formulation assumes that the CoM is located at the base frame origin (its location being therefore constant in this frame), which is a good approximation since most of the robot mass is located in its base.
It also assumes a constant locked inertia tensor to ensure  consistency when mixing orientation and angular momentum control. 
Force allocation at each foot in contact is computed at each control cycle by solving the following quadratic program
\vspace{-0.1cm}
\begin{align}
\qquad&\min_{\mathbf{F}_i, \boldsymbol{\eta}, \zeta_1, \zeta_2} \sum_i \mathbf{F}_i^2 + \alpha(\boldsymbol{\eta}^2 + \zeta_1^2 + \zeta_2^2) \nonumber \\
\textrm{s.t.}\qquad & \mathbf{W}_{CoM} = \sum_{i\in \mathcal{C}} \left(\begin{matrix} \mathbf{F}_i \\ \mathbf{r}_i \times \mathbf{F}_i \end{matrix} \right) + \boldsymbol{\eta} \nonumber\\
& \begin{matrix} F_{i,x} < \mu F_{i,z} + \zeta_1,\ \ F_{i,y} < \mu F_{i,z} + \zeta_2 \end{matrix} \qquad \forall i\in \mathcal{C} \nonumber
\end{align}
where $\mathcal{C}$ is the set of indexes for the feet in contact with the ground (contact activation is decided using both the plan and the contact-sensor feedback), $\boldsymbol{\eta}$, $\zeta_1$ and $\zeta_2$ are slack variables that ensure that the QP is always feasible, $\alpha$ is a large weight, $\mathbf{r}_i$ is the vector
from foot $i$ to CoM, $\mu$ is the friction coefficient and $z$ is the direction orthogonal to the ground.
Once an optimal foot force allocation is found, the actuation torques are computed as
\vspace{-0.1cm}
\begin{equation}
    \boldsymbol{\tau_i} = \mathbf{J}_{i,a}^T  \left( \mathbf{F}_i +
    \mathbf{K} (\mathbf{l}_i^{ref} - \mathbf{l}_i) + \mathbf{D} (\dot{\mathbf{l}}_i^{ref} - \dot{\mathbf{l}}_i) \right) \nonumber
\end{equation}
where $\mathbf{J}_{i,a}$ is the actuated part of the foot Jacobian, 
$\mathbf{l}$ is the vector between the foot and the base origin, and the index $i$ corresponds to each leg. The control law combines a desired force and a low-impedance leg-length controller.
%
\subsubsection{Solo motion capabilities}
Solo's leg joints are multi-turn capable, up to three turns in each direction.
Figure~\ref{fig:dexterity} illustrates how Solo can exploit these capabilities
in various situations.
%
Solo's knee joints can be bent in both directions, and configure the robot into `X', `O'-knee postures, or forward- and backward `C' postures as required (Fig.\,\ref{fig:dexterity}a). 
For example, it bends its knee joint backward to reach the obstacle from the top in Fig.\,\ref{fig:dexterity}b.
We designed a simple motion sequence (Fig.\,~\ref{fig:dexterity}c), allowing Solo to stand up after a turn-over. We note that
the three demonstrated behaviors cannot be kinematically achieved by quadruped animals.
\subsubsection{Tracking kino-dynamic plans}
We first tested the controller in different balancing scenarios that can be seen
in the accompanying video. The results show that the robot is capable of balancing
on moving platforms without any knowledge of the environment.
We also computed a jumping motion, as well as both slow and fast walks. Results show that the
robot can follow the desired plans. While the plans in these experiments are relatively long (e.g. $\SI{15}{s}$ for the walking experiments), no re-planning is required for the robot
to achieve the task.
During the slow-walk motion, we added a seesaw obstacle that is not taken into
account in the planner nor the controller. The robot is able to traverse the terrain without  any problems.
These results suggest that the controller is robust to uncertain environments and can adequately stabilize long motion plans.
Moreover, this demonstrates that the plans generated by the kino-dynamic optimizer
can be transferred to a real robot despite the difference between the dynamic model of
the robot and the real system.
We also executed a vertical jump, reaching a base height of \SI{65}{cm} (Fig. \ref{fig: solo quadruped}), and landing without damage.
%
%

%

\section{Discussion}
\subsubsection{Design choices}
Designing a low-weight quadruped robot while maintaining effective impedance and force control capabilities required us to trade-off several design features. 
We  designed  our  legged  robot  architecture  for  proprioceptive force-control using BLDC motors as they lead to high performance and low weight actuators \cite{seok_actuator_2012}, with sufficiently low gear ratio (9:1) to achieve proprioceptive actuation~\cite{ding_design_2017,ramos_facilitating_nodate}. 
Such actuators do not require dedicated force sensors
as joint torque is directly estimated from 
motor phase current measurement~\cite{wensing_proprioceptive_2017}.
%
Low-geared actuators have the advantage of low friction and minimal stage losses, while fewer and smaller parts in the gear train reduce the losses from reflected inertia under oscillating loads~\cite{roos_optimal_2006}. 
Choosing to utilize low-cost hobbyist BLDC motors introduces torque ripples at very low speed. It has not been a problem yet during normal locomotion operation but might necessitate active compensation for slow precision tasks. 

%

Legged robots experience high peak torques from impacts, imposing high dynamic loads on the gear train components.  The small contact surface of single-point contact spur-gears are not well suited to such loads, and are relatively rare in jumping robots unless in combination with mechanical compliance mounted in the leg-length~\cite{meyer_passive_2006,sato2015development} or leg-angle direction~\cite{eckert_towards_2019}. 
While planetary gears share loading among multiple teeth-~\cite{wensing_proprioceptive_2017}, cable-~\cite{kitano_titan-xiii:_2016,hwangbo2018cable}, belt-~\cite{ramos_facilitating_nodate}, and chain-driven~\cite{hutter2011scarleth,katz2019mini} systems typically exhibit high robustness against external peak torque and can transmit power over a larger distance with low reflected inertia.
We decided for a low weight, dual-stage timing belt transmission with 9:1 reduction capable of sustaining high impacts.

Proprioceptive actuation does not require additional sensing for locomotion on flat and level ground; however, on unknown terrain it becomes difficult to reliably estimate rapid, low-force contacts.
We chose to combine proprioceptive control with robust sensing using a distally mounted touch sensor.
Only a few sensing principles remain functional at the harsh impact conditions experienced during high-speed leg touchdowns.
Peak forces can exceed two times bodyweight when exerted onto a single foot~\cite{wensing_proprioceptive_2017}.
Lightweight force-sensing based on piezoresistive or optical sensing of deflecting elastic material has been demonstrated previously~\cite{ramos_facilitating_nodate,kim_note:_2013}.
Other designs measure the deflection of rubber-like materials through embedded magnets~\cite{ananthanarayanan2012compact} or measure impacts with inertial measurement units~\cite{kaslin2018towards}.
These sensing concepts are relatively complex.
Here we propose a simple and inexpensive design based on a spring-loaded aperture, similar to the principle implemented by Hsieh~\cite{hsieh_three-axis_2006}.


\subsubsection{Impedance control capabilities}
We measured torque at the motor through current measurement.
The resulting output torque at the end effector differed due to gear losses, belt and leg structural flexibility, and inertial losses. 
Nevertheless, our experimental results show that very good impedance control is possible.
%
The systematic characterization of leg stiffness suggests that the actuator module can serve as a basis to construct high performance force-controlled robots. As a comparison, reported human leg stiffness values while running~\cite{farley_leg_1996} ranges from $k \seq \SI{7}{kN/m}$ to \SI{16.3}{kN/m}. 
For a \SI{75}{kg}, \SI{1}{m} leg length human, this translates into a dimensionless leg stiffness, $\tilde{k} \seq  k \cdot l_0/(mg)$~\cite{rummel_stable_2008}, between  $\tilde{k} \seq 10$ and $\num{22}$.
In our 2-DOF leg experiments we measured \SI{266}{N/m} stiffness, corresponding to a dimensionless leg stiffness of $10.8$, putting the capabilities of the robot within a range comparable to human leg stiffness.
Comparison with other quadruped robots is difficult as impedance or force
control performance is seldom characterised.
From the reported characterization of the \SI{10}{kg}, hydraulically driven HyQ leg \cite[Fig 12]{semini_towards_2015}, we 
 estimate a dimensionless HyQ leg stiffness of $\tilde{k} \seq \SI{5250}{N/m}  \cdot  \SI{0.3}{m} / (\SI{10}{kg} \cdot  \SI{9.81}{m/s^2}) \seq 16$, slightly higher than the dimensionless stiffness of Solo.

%





\subsubsection{Open Source and Outreach}
Mechanical and electrical hardware blueprints and software required for this project are open-source under the BSD-3-clause license \cite{opensourcelink} and the robots can be easily reproduced and improved by other laboratories. At the moment,
three other laboratories are in the process of producing their own copy of the quadruped and a fully wireless 12-DOF quadruped with adduction/abduction degree of freedom at the hip joint is currently under construction.
The actuator module is inexpensive, and the full 8-DOF quadruped was built for approximately \SI{4000}{\euro} of material cost.
The low weight and simplicity of the robot allow for easy transportation and safe operation, significantly simplifying experimental environments. 
We have demonstrated legs assembled from the same actuator module but
other configurations are also possible. 
For example, multiple legs modules can be used as manipulators when reconfigured into a large hand-like structure. 
The platform can also be used as an educational tool. For example, the leg was used to teach robotics to high school interns at NYU.

\section{Conclusion}
We presented an open-source low-cost actuator module and a foot contact sensor used to build torque-controlled legged robots. 
We developed the system's hardware, electronics, and firmware/software to support legged robot locomotion research with a rugged and durable, low-weight robot that can be handled safely by a single researcher. 
Experiments show the robots' capabilities in generating very dynamic motions with excellent impedance regulation characteristics.
We introduced a simple torque-controller capable of regulating motions generated with full-body kino-dynamic optimizer.
We anticipate that this open-source project will benefit the robotics community by lowering the barrier to entry and lead to further extensions of the robots.

\bibliographystyle{IEEEtran}
\bibliography{literature2}

\end{document}